\documentclass[sigconf]{acmart}

\AtBeginDocument{%
  }

\copyrightyear{2024} 
\acmYear{2024} 
\setcopyright{rightsretained} 
\acmConference[CVMP '24]{European Conference on Visual Media Production}{November 18--19, 2024}{London, United Kingdom}
\acmBooktitle{European Conference on Visual Media Production (CVMP '24), November 18--19, 2024, London, United Kingdom}\acmDOI{10.1145/3697294.3697309}
\acmISBN{979-8-4007-1281-4/24/11}

\begin{document}

\title{Multi-Resolution Generative Modeling of Human Motion from Limited Data}

\author{David Eduardo Moreno-Villamar\'in}
\affiliation{%
  \institution{Fraunhofer Heinrich Hertz Institute, HHI}
  \city{Berlin}
  \country{Germany}
}
\email{david.moreno@hhi.fraunhofer.de}
\orcid{0002-6807-4845}

\author{Anna Hilsmann}
\affiliation{%
  \institution{Fraunhofer Heinrich Hertz Institute, HHI}
  \city{Berlin}
  \country{Germany}
}
\email{anna.hilsmann@hhi.fraunhofer.de}
\orcid{0002-2086-0951}

\author{Peter Eisert}
\affiliation{%
  \institution{Fraunhofer Heinrich Hertz Institute, HHI}
  \city{Berlin}
  \country{Germany}
}
\affiliation{%
  \institution{Humboldt University of Berlin}
  \city{Berlin}
  \country{Germany}
}
\email{peter.eisert@hhi.fraunhofer.de}
\orcid{0001-8378-4805}

% \renewcommand{\shortauthors}{Moreno-Villamarin et al.}

\begin{abstract}
    We present a generative model that learns to synthesize human motion from limited training sequences. Our framework provides conditional generation and blending across multiple temporal resolutions. The model adeptly captures human motion patterns by integrating skeletal convolution layers and a multi-scale architecture. Our model contains a set of generative and adversarial networks, along with embedding modules, each tailored for generating motions at specific frame rates while exerting control over their content and details. Notably, our approach also extends to the synthesis of co-speech gestures, demonstrating its ability to generate synchronized gestures from speech inputs, even with limited paired data. Through direct synthesis of SMPL pose parameters, our approach avoids test-time adjustments to fit human body meshes. Experimental results showcase our model's ability to achieve extensive coverage of training examples, while generating diverse motions, as indicated by local and global diversity metrics. % motionvation/importance of style is not well explained
    % so also modify style definition/control here
    % Repeated references to Ganimator and direct comparison only with this method. Makes the reader to think that is a "extension paper"
\end{abstract}


\begin{CCSXML}
<ccs2012>
   <concept>
       <concept_id>10010147.10010371.10010352.10010380</concept_id>
       <concept_desc>Computing methodologies~Motion processing</concept_desc>
       <concept_significance>500</concept_significance>
       </concept>
 </ccs2012>
\end{CCSXML}

\ccsdesc[500]{Computing methodologies~Motion processing}

% \ccsdesc[300]{Neural networks}
% \ccsdesc{Neural networks}

\keywords{Generative models, machine learning, motion synthesis, gestures}
\begin{teaserfigure}
  \includegraphics[width=\textwidth]{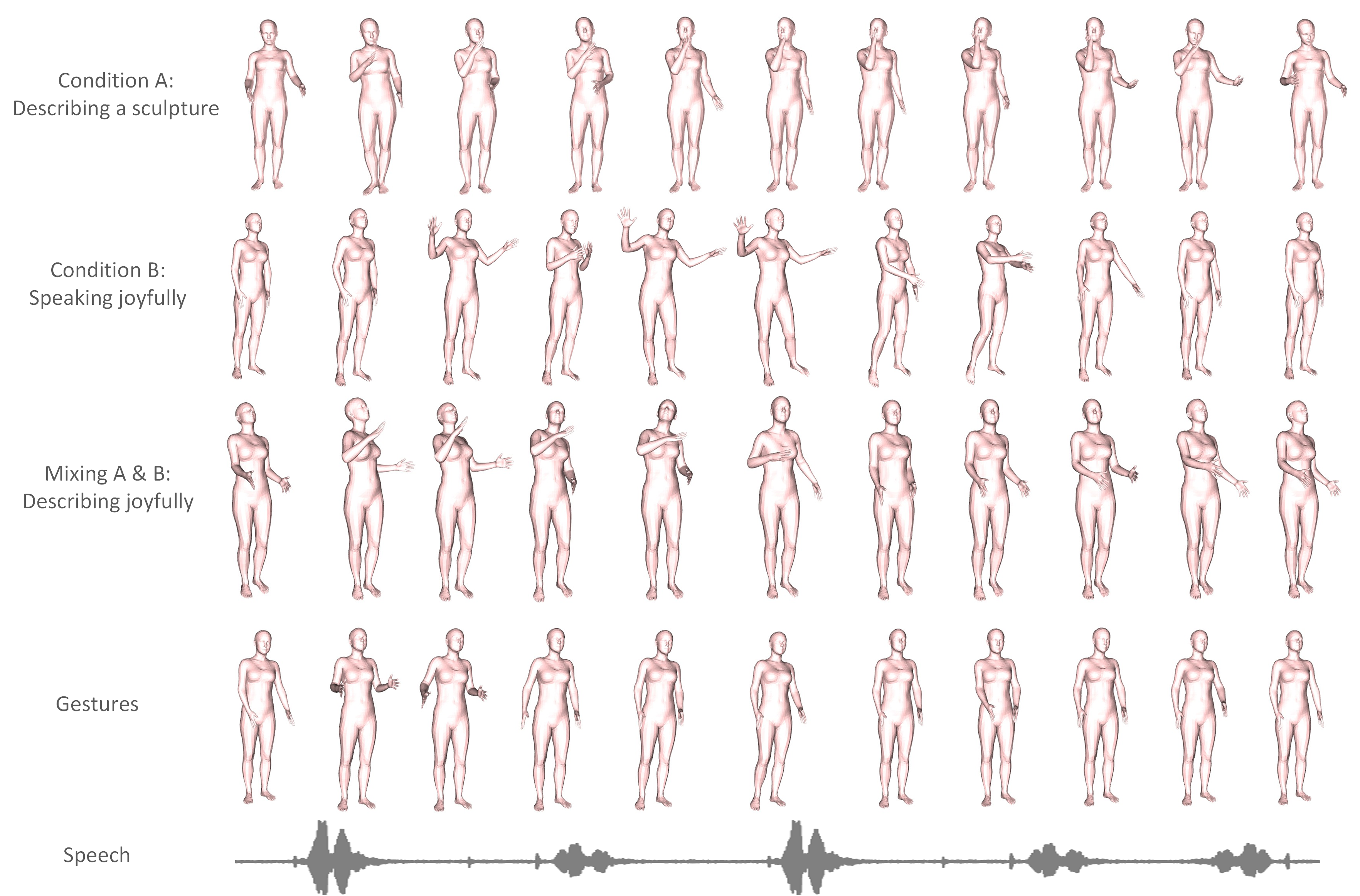}
  \caption{We present our motion synthesis framework for generating human motion based on single (first and second row) or multiple conditions (third row) across varying temporal scales, even when trained on limited data. Additionally, we demonstrate the framework's application in synthesizing co-speech gestures with similarly constrained data.}
  \Description{Rendered avatars showing the model application.}
  \label{fig:teaser}
\end{teaserfigure}


\maketitle

\section{Introduction}
\label{sec:intro}

The realm of modeling and synthesizing human motion is a thoroughly investigated domain within computer graphics, driven by diverse applications such as virtual avatar creation, entertainment, and social robots. Despite notable advancements towards natural and realistic motion synthesis \cite{alexanderson2023listen,petrovich22temos}, the persisting challenge lies in identifying a representation enabling humans to perceive fine-grain motion patterns, which is typically lost in more abstracted forms \cite{ng2024audio2photoreal}. This reliance introduces constraints on the accessibility of training data, the evolution of motion synthesis models, and their practical applicability.

For instance, substantial efforts have been devoted to developing  models of realistic human bodies capable of representing diverse body shapes and emulating  soft-tissue motions \cite{loper2015smpl, MANO:SIGGRAPHASIA:2017,pavlakos2019smplx, Osman2020}. However, capturing and fitting volumetric or motion capture (MoCap) data to these body models is costly and tedious, resulting in a lack of volume and diversity of motion in existing datasets for different synthesis tasks.

Motivated by these challenges, we present a generative model that learns to synthesize novel motions from limited motion sequences. We draw inspiration from the work of Li et al.~\cite{li2022ganimator} while incorporating control signals at different time scales to enhance the flexibility and adaptability of the generated motions. For example, our model can generate a motion sequence of a person speaking, where the motion details are influenced by emotions present in other sequences, such as anger or joy.  Furthermore, we design our model to synthesize sequences of pose parameters of SMPL \cite{loper2015smpl}, which is standard in current 3D human animation frameworks given its capability to represent realistic and diverse body shapes.

% Paper structure
This paper is structured as follows: In Section \ref{sec:rel}, we provide an overview of related work pertinent to our task. Section \ref{sec:method} presents our proposed method, outlining its key components and functionalities. The experiments conducted to evaluate our method are detailed in Section \ref{sec:exper}. Section \ref{sec:appl} shows the application of our method to gesture synthesis. Finally, Section \ref{sec:concl} summarizes our findings and draws conclusions from our work.

\section{Related Work}
\label{sec:rel}

% details of the blocks, such as the parameters, should be clearly defined and explained directly in the figures even they have been explained in text. 

We present a concise overview of pertinent literature on human motion synthesis, delving into various works that concentrate on motion generation, with a particular emphasis on human animation.

Early applications of deep learning in motion synthesis \cite{Holden2016, Fragkiadaki2015} demonstrated the efficacy of recurrent neural networks in capturing human dynamics to model and synthesize motion. Subsequent research efforts \cite{Aksan2019, aberman2020skeleton} have highlighted the improved outcomes achieved by integrating skeletal structure information into network architectures. Henter et al.~\cite{henter2020moglow} introduced a probabilistic, generative, and controllable motion model based on normalizing flows \cite{Rezende2015}. Recent works have addressed various motion synthesis tasks, wherein generation is conditioned on specific control inputs. Examples include synthesizing actions based on labels, trajectories, or textual descriptions \cite{henter2020moglow, Jang2020, petrovich2021actor, petrovich22temos, petrovich2024multi}, generating dance moves conditioned on audio cues \cite{Huang2020, Yuan2020, lee2020dancemotions}, and producing co-speech gestures in response to audio, text, or a combination of both \cite{Ferstl2019, Yoon2019, Kucherenko2019analyzing, Yoon2020, Kucherenko2020, Habibie_IVA2021, xu2022freeform, ng2024audio2photoreal}. 

The effectiveness of deep learning methods for motion synthesis is attributed to the accessibility of extensive motion capture datasets. Nevertheless, procuring such a substantial amount of data is expensive,  and as mentioned in Section \ref{sec:intro}, introduces additional challenges when integrated with different human animation models, since captured skeletons may differ in their structure and number of joints. Li et al.~\cite{li2022ganimator} address these data availability obstacles and present a patch GAN-based method that learns to synthesize varied motions from a single motion capture sequence. However, their approach offers limited control over the synthesis process (e.g. user-defined root joint movement). % maybe i can think of how to expand this a bit
%, as  a set of used-defined constraints of a subset of joints (e.g. root joint movement) or filling in details of a user-provided coarse sequence with the details of a sequence the network was trained on.
%However, their approach does not offer control over the synthesized motion beyond constraining over a subset of joints, for instance, the trajectory of a character. % instead of however: "although their approach offers style transfer and motion mixing"... or that motion mixing doesn't allow to control where, when, how to include both motions,and style transfer requires an already provided/generated motion sequence.
% i can also add some sentences here about Li's limitations: yes, it generates from single sequence and allows to have some variability from it, but does not offer control beyond conditioning on some joints, or more specifically, the global movement of the character.

In contrast, we directly model and generate human motion on SMPL pose parameters, which are consistent with the human anatomy and mitigate expensive test-time optimizations to fit a human body mesh. Furthermore, our approach is able to synthesize diverse motion patterns from limited motion sequences,
while offering control over different time-scales in terms of emotions, actions, or other signals (e.g. speech). This is done by learning a set of embeddings from multi-class annotations or features extracted from a control signal (e.g. speech). The embeddings are used by a feature-wise linear modulation (FiLM) layer that integrates the corresponding information into intermediate motion features. We show that by learning such embeddings, we are able control the output motion without compromising the coverage or diversity of the generated motion. % i want to check the methods mentioned in genea to post-processed synthesized gestures.

%while offering multi-resolution stylistic control by learning style embeddings from the training data.
% we, in theory, need more than a couple of sequences, but we offer that we can exert some control on the output by having annotated sequences related to emotions or actions, and we can learn how to control and combine them thanks to the multiresolution approach and the feature modulation network. Also add that in this case multi-resolution is time resolution./ time dimension

% to further explain, i could add a sort of generalization as input to the gan. like im using action/emotion labels, but that in the end it can be just a label describing a sequence and the gan learns to generate in such style?
% how are these obtained from the data? we just annotate the sequences with one-hot vectors
% motivation/importance of style is not well explained, maybe we could add that ganimator lacks controllability apart from trajectory control? and could use other inputs?
% maybe also change the term 'style' from something different. i think i only got it from being inspired by stylegan

\section{Method}
\label{sec:method}

This section presents our human motion synthesis framework with multi-resolution control. We begin by introducing the notations regarding the employed motion representation, followed by an overview of our network architecture. Finally, we describe details about each network component.
\begin{figure*}
    \centering
    \includegraphics[width=\linewidth]{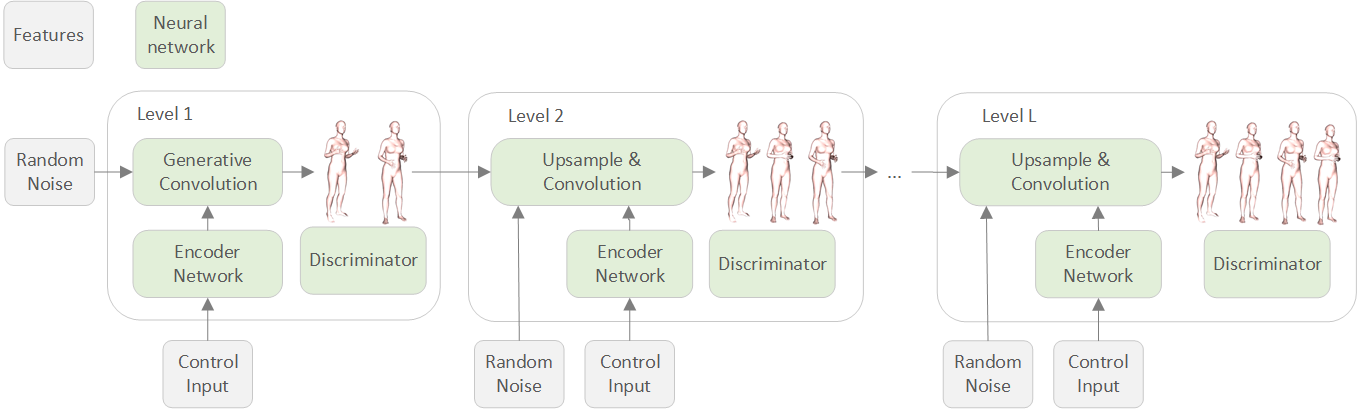}
    \caption{Overview of our motion synthesis architecture. The network generates an initial sequence of keyframes from a random noise vector and a control input. Subsequently, the sequence undergoes progressive upsampling in the temporal dimension until reaching the original resolution of the training sequence or sequences. At each level, encoders integrate control information into the motion, enabling the combination of different control signals at various temporal resolution levels during inference. We train our framework in an adversarial manner, employing a set of discriminators to evaluate the synthesized motions at each level.}
    \Description{.}
    \label{fig:arch}
\end{figure*}

\subsection{Motion Representation}
\label{sec:repr}

% we base our approach on the human body representation SMPL. it consists of a differentiable function that takes as parameters pose, decribing the rotations of a skeleton, and shape, containing information about the blend shapes of the individual. The function takes as input these parameters, and outputs the vertices of the posed mesh. ksakld
% , as depicted in \autoref{fig:method}
Our objective is to learn a generative model able to animate a human's mesh by synthesizing motion according to a control signal. We produce a mesh for each frame $t$ by leveraging the SMPL body model \cite{loper2015smpl}, which takes as input pose parameters $\theta^t$ and shape parameters $\beta$. The pose parameters are $Q$-dimensional rotation features for $K$ joints of the SMPL skeleton, while the shape parameters are subject-specific and control the overall body shape. The SMPL model functions as a differentiable mapping $M(\beta,\theta)$, that translates these parameters into mesh vertices.

To address foot sliding artifacts, we also model foot contacts. We use binary labels to indicate whether each foot is in contact with the ground. We concatenate to the feature axis $C$ the binary contact labels $c \in \{0,1\}^C$, which correspond to the contact status of the foot vertices $\mathcal{F}$ in the mesh. For the SMPL model, we use $\mathcal{F}=\{$left big toe, left small toe, left heel, right big toe, right small toe, right heel$\}$. For each foot vertex $j\in \mathcal{F}$ and frame $t\in\{1,...,T\}$, we compute the $c^{tj}$ label as:

\begin{equation}
    c^{tj} = \mathbb{I}\left[\|V^j(\beta,\theta^t)\|_2 < \epsilon\right],
\end{equation}

where $\|V^j(\beta,\theta^t)\|_2$ represents the magnitude of the velocity of vertex $j$ at frame $t$, derived from the SMPL model $M$. The function $\mathbb{I}[\cdot]$ is an indicator function that evaluates to $1$ if the condition inside the brackets is met and $0$ otherwise.

In our framework, we define motion as a sequence of $T$ vectors $x^{1:T}$, where $|x^t| = KQ + C +3$. Each vector encompasses SMPL pose parameters $\theta^t$, along with $3$ dimensions for the root joint displacement, and concatenate foot contact labels along $C$ channels. We express each rotation as 6D features, as proposed by Zhou et al. \cite{zhou2019continuity}, known for achieving optimal performance in deep learning frameworks \cite{bregier2021deep}.
% i should distinguish pose/motion features in our framework from SMPL pose features.

We denote the space of motion features as $\mathcal{M}_\Theta\equiv \mathbb{R}^{T\times (KQ+C+3)}$ for simplicity. Furthermore, we represent motion features as  $\Theta \equiv [x^{1:T}] \in \mathcal{M}_\Theta$, and their corresponding temporally downsampled versions as $\Theta_i \in \mathcal{M}_{\Theta_i}$, while we denote a set with $N$ samples as $\{\Theta^k\}_{k=1}^N$.

Additionally, we introduce a signal $s$ to guide our motion generation model. Our experiments explore scenarios where $s$ functions as a multi-class label, and speech features extracted from an audio signal, as detailed in Sections \ref{sec:exper} and \ref{sec:appl}. This design choice enables us to control the generated motion across different temporal resolution levels. For example, we can assign a condition $s^{a}$ to guide the generation of a temporally coarse motion sequence and another signal $s^b$ for incorporating finer motion details. This approach grants the model the capability to control the synthesized motion, beyond simply mixing of the training sequences.
% maybe add here more about speech, labels, or more how it can be anything but we have experiments with these

\subsection{Model Architecture}

% i need to add somewhere what the control signal actually is, maybe at end of section?

Our motion synthesis architecture, depicted in \autoref{fig:arch}, is able to guide the generation of content and details disjointedly. We achieve this by applying a set of encoder and modulation layers across multiple time scales. Both encoder and modulation layers integrate a control signal $s$ into intermediate motion features at a specific temporal resolution, which enables increased functionality and adaptability.

We draw inspiration from multi-scale frameworks \cite{li2022ganimator, karras2018progressive, shaham2019singan, shocher2019ingan} and construct our architecture with a set of $L$ generative adversarial networks (GANs) \cite{goodfellow2014generative}. Each GAN is responsible for synthesizing motion at a specific temporal resolution $i$. Within this set, we represent generators and discriminators as $\{G_i\}_{i=1}^L$ and $\{D_i\}_{i=1}^L$, respectively. An initial generator $G_1$ maps a random noise vector $z_1 \in \mathcal{M}_{\Theta_1}$ and a control signal $s$ to generate a temporally coarse motion sequence

\begin{align}
    \hat{\Theta}_1 &= G_1(z_1, s)\\
    G_1(z_1, s) &= G_1^*(S_1(z_1, s)),
\end{align}

where $S_1$ is a network that encodes $s$ to modulate $z_1$, and $G_1^*$ is a convolutional network that generates a temporally coarse motion sequence $\hat{\Theta}_1 \in \mathcal{M}_{\Theta_1}$. We progressively generate finer sequences for levels $2\leq i\leq L$:

\begin{align}
    \hat{\Theta}_i &= G_i(\hat{\Theta}_{i-1}, s, z_i)\\
    G_i(\hat{\Theta}_{i-1}, s_i, z_i) &= G_i^*(S_i(\hat{\Theta}_{i-1}, s, z_i)).
\end{align}
In each level, a sequence $\hat{\Theta}_{i-1}$ from the previous level is temporally upsampled by a fixed scaling factor $F>1$, and added to a noise vector $z_i$. Similar to the first level, $S_i$ modulates the upsampled noisy sequence with $s$, while $G_i^*$ takes the modulated features to generate a sequence $\hat{\Theta}_i$ in the $i$-th time scale. This process repeats until $G_L$ generates the temporally finest output sequence $\hat{\Theta}_L \in \mathcal{M}_{\Theta_L}$ according to the control input $s$. We carry out experiments where $s$ is a one-hot vector denoting a motion sequence related to an action or emotion, and where $s$ is a signal of speech features, as described in Sections \ref{sec:exper} and \ref{sec:appl}. 

For each step, $z_i$ follows a i.i.d. normal distribution $\sim \mathcal{N}(0, \sigma_i)$ along the temporal axis while being shared along the channel axis. In \cite{li2022ganimator}, the authors found that $\sigma_i$ is correlated with the high frequency details generated by each $G_i$, and define it at each level as an error between an upsampled $\Theta_{i-1}$ and $\Theta_i$. In our experiments, we select $F=4/3$ and $L=8$ for a label-based approach and $L=10$ for speech-guided synthesis.

\subsection{Network Components}

The generator and discriminator networks $G_i$ and $D_i$ in our architecture follow similar structures as explained in \cite{li2022ganimator}. These generators, contain fully convolutional neural networks $g_i^*(\cdot)$ that present skeleton-aware convolution layers \cite{aberman2020skeleton}. We maintain a residual structure for generators ($2\leq i \leq L$) as the primary role of these networks is to add missing high-frequency details:

\begin{align}
    G_i^*(\hat{\Theta}_{i-1}, s_i, z_i) &= g_i(\hat{\Theta}_{i-1}, s_i, z_i) + \uparrow \hat{\Theta}_{i-1}\\
    g_i(\hat{\Theta}_{i-1}, s_i, z_i) &= g_i^*(S_i(\uparrow \hat{\Theta}_{i-1} + z_i, s_i)).
\end{align}

\begin{figure}[htb]
    \centering
    \includegraphics[width=\linewidth]{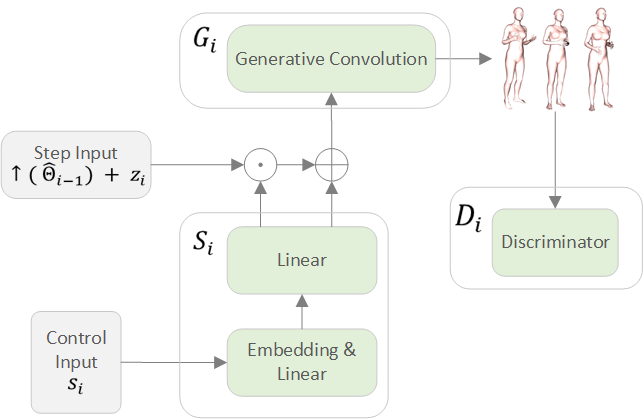}
    \caption{Generator Structure: At each step $i$, a neural network $S_i$ embeds a control input $s$ into parameters $\gamma_i$ and $\delta_i$ that modulate the current step's input via Feature-wise Linear Modulation (FiLM) \cite{perez2018film}. The current step's input is a noise vector $z_i$ added to an upsampled version of the previous step's result $\hat{\Theta}_{i-1}$. The generator's role is to predict missing high-frequency details.}
    \Description{.}
    \label{fig:gen_step}
\end{figure}

Each network $S_i$ firstly embeds the control signal $s_i$ to modulate the input to the current level through Feature-wise Linear Modulation (FiLM) \cite{perez2018film}. More specifically, we apply a fully connected layer to embed $s_i$ into an intermediate representation $y_i$. We use FiLM to learn functions $f$ and $h$ which provide $\gamma_i$ and $\delta_i$ as a function of features $y_i$:

\begin{align}
    \gamma_i &= f(y_i)\\
    \delta_i &= h(y_i),
\end{align}

where $\gamma_i$ and $\delta_i$ modulate upsampled motion features $\uparrow(\hat{\Theta}_{i-1})$ and variation source $z_i$ as:

\begin{equation}
    \text{FiLM}(\hat{\Theta}_{i-1}|\gamma_i,\delta_i) = \gamma_i \left(\uparrow(\hat{\Theta}_{i-1}) + z_i\right) + \delta_i,
\end{equation}

where $\uparrow(\cdot)$ is a linear upsampler with scale value $F>1$. \autoref{fig:gen_step} illustrates our network at each step. % again i can explain here what is the s input. which is a label or features from something.
% also add something about discriminators

To prevent mode collapse and overfitting when training our set of GANs, we employ a Patch-GAN discriminator \cite{isola2017image} as suggested in \cite{li2022ganimator}. This discriminator structure limits the receptive field by evaluating short temporal patches of the motion sequence independently, while assigning a confidence value to each. The final output is the average of these per-patch values, which promotes the generation of more varied and realistic motion by ensuring that the discriminator focuses on local coherence and detail.

\subsection{Training}

% details of the blocks, such as the parameters, should be clearly defined and explained directly in the figures even they have been explained in text. 

We optimize the network parameters for each resolution step $i$ by minimizing the following loss function

\begin{equation}
\label{eq:loss}
    \mathcal{L} = \lambda_{\text{adv}} \mathcal{L}_{\text{adv}} + \lambda_{\text{con}}\mathcal{L}_{\text{con}} + \lambda_{\text{rec}} \mathcal{L}_{\text{rec}} + \lambda_{\text{smooth}}\mathcal{L}_{\text{smooth}},
\end{equation}

where $\mathcal{L}_{\text{adv}}$ is the adversarial loss that guides the training of our GANs, $\mathcal{L}_{\text{con}}$ is the contact consistency loss to handle foot contact with the ground, $\mathcal{L}_{\text{rec}}$ is the reconstruction loss, and $\mathcal{L}_{\text{smooth}}$ ensures smooth transitions in the generated motion. To enhance robustness and quality, we train the model block by block, grouping every two consecutive levels \cite{li2022ganimator, hinz2021improved}.

The adversarial loss is computed using the WGAN-GP formulation \cite{arjovsky2017wgan, gulrajani2017improved}, given by:

\begin{align}
\label{eq:wgan}
    \mathcal{L}_{\text{adv}} & = \mathbb{E}_{\hat{\Theta}_i\sim\mathbb{P}_{g_i}}[D_i(\hat{\Theta}_i)] - D_i(\Theta_i)\\
    &+ \lambda_{\text{gp}}\mathbb{E}_{\tilde{\Theta}_i\sim\mathbb{P}_{\tilde{g}_i}}\left[\left(\|\nabla D_i(\tilde{\Theta}_i)\|_2-1\right)^2\right],
\end{align}

where $\mathbb{P}_{g_i}$ represents the distribution of generated samples at level $i$, $\mathbb{P}_{\tilde{g}_i}$ is the distribution of interpolated samples between real and generated data, with $\tilde{\Theta}_i=\lambda\hat{\Theta}_i+(1-\lambda)\Theta_i$ and $\lambda\sim\text{Uniform}(0,1)$, and $\lambda_{\text{gp}}$ is the weight of the gradient penalty term. The gradient penalty enforces Lipschitz continuity, to accurately approximate the Wasserstein distance between real and generated distributions \cite{gulrajani2017improved}.

To address foot sliding artifacts, we incorporate a contact consistency loss $\mathcal{L}_{\text{con}}$. Our model predicts foot contact labels from foot vertices, and these predictions are refined via a post-processing step using inverse kinematics. This loss encourages consistency between predicted contact labels and foot velocity:

\begin{equation}
    \mathcal{L}_{\text{con}} = \frac{1}{T|\mathcal{F}|}\sum_{j\in\mathcal{F}}\sum_{t=1}^T\left\|V(\beta, \theta^t)^j\right\|_2^2\cdot \text{sig}(c^{tj}),
\end{equation}
where $\text{sig}(x)=1/[1+\exp(5+-10x)]$ is the transformed sigmoid function. This term ensures that either the contact label $c^{tj}$ or the foot velocity is minimized, resulting in more natural motion and preventing inconsistent foot behavior during post-processing \cite{li2022ganimator}.

We employ a reconstruction loss term to ensure the network synthesizes variations of all temporal patches of the training sequences. We achieve this by forcing the network to replicate an input motion using pre-defined noise signals $\{z_i^*\}_{i=1}^L$ and their corresponding control signal $s^*$. In particular, $ G_i$ should reconstruct all examples $\Theta_i^k$ at level $i$ through the following loss:

\begin{equation}
    \label{eq:recloss}
    \mathcal{L}_{\text{rec}} = \frac{1}{N}\sum_{k=1}^N\left\|G_i(\Theta_{i-1}^k,s_i^{k*},z_i^{k*})-\Theta_i^k\right\|_1.
\end{equation}

During reconstruction, we expect no noise influence beyond the coarsest temporal level \cite{li2022ganimator}, so we set $z_1^*$ as the pre-generated noise for the coarsest level and $z_i^*=0$ for $i> 1$.

To ensure smooth transitions between frames in the finer temporal resolution levels, we include a smoothness loss term defined as:

\begin{equation}
    \mathcal{L}_{\text{smooth}} = \frac{1}{T}\sum_{t=1}^T\left\|p^t - \frac{p^{t-1}+p^t+p^{t+1}}{3}\right\|_2,
\end{equation}

where $p^t$ represents the 3D joint locations at frame $t$. These joint locations are derived from the SMPL pose parameters $\theta^t$ using a joint regressor, which computes the positions from the mesh based on the body shape. By minimizing the difference between each frame's joint locations and the average of its neighboring frames, this term encourages temporal coherence, resulting in smoother human motion.

\begin{table*}[htb]
\caption{Quantitative comparison.}
\centering
\begin{tabular}{lcccc}
\toprule
 & Sequence & Coverage $\uparrow$& Global Diversity $\uparrow$ & Local Diversity $\uparrow$\\
\midrule
GANimator \cite{li2022ganimator} & A & $98.7\%$ & $0.93$ & $0.90$ \\
& B & $98.0\%$ & $0.95$ & $0.93$\\
& C & $88.1\%$ & $1.15$ & $1.10$\\
& D & $90.2\%$ & $1.02$ & $1.00$\\
& E & $97.2\%$ & $1.01$ & $0.98$\\
\midrule
Ours & A & $100\%$ & $1.04$ & $1.06$ \\
& B & $99.9\%$ & $1.03$ & $1.00$\\
& C & $99.0\%$ & $1.22$ & $1.18$\\
& D & $99.4\%$ & $1.08$ & $1.06$\\
& E & $99.4\%$ & $1.06$ & $1.03$\\
\bottomrule
\end{tabular}
\label{tab:comparison}
\end{table*}

\section{Experimental Setup}
\label{sec:exper}

%some summary, task description, and such.
% \subsection{Tasks}

To evaluate the effectiveness of our conditional framework, we test the task of conditional motion generation. This task focuses on synthesizing motion sequences using simple conditions. Specifically, we aim to generate motion driven by a set of instance labels, each represented as a one-hot vector. These labels serve as control signals, or inputs to our set of encoder networks $\{S_i\}_{i=1}^L$, which transform this categorical information into FiLM parameters. These FiLM parameters, in turn, modulate the motion features, allowing us to generate sequences that adhere to the specified conditions.

In the following sections, we provide detailed descriptions of how the data is prepared and utilized.

\begin{figure*}
    \centering
    \includegraphics[width=\linewidth]{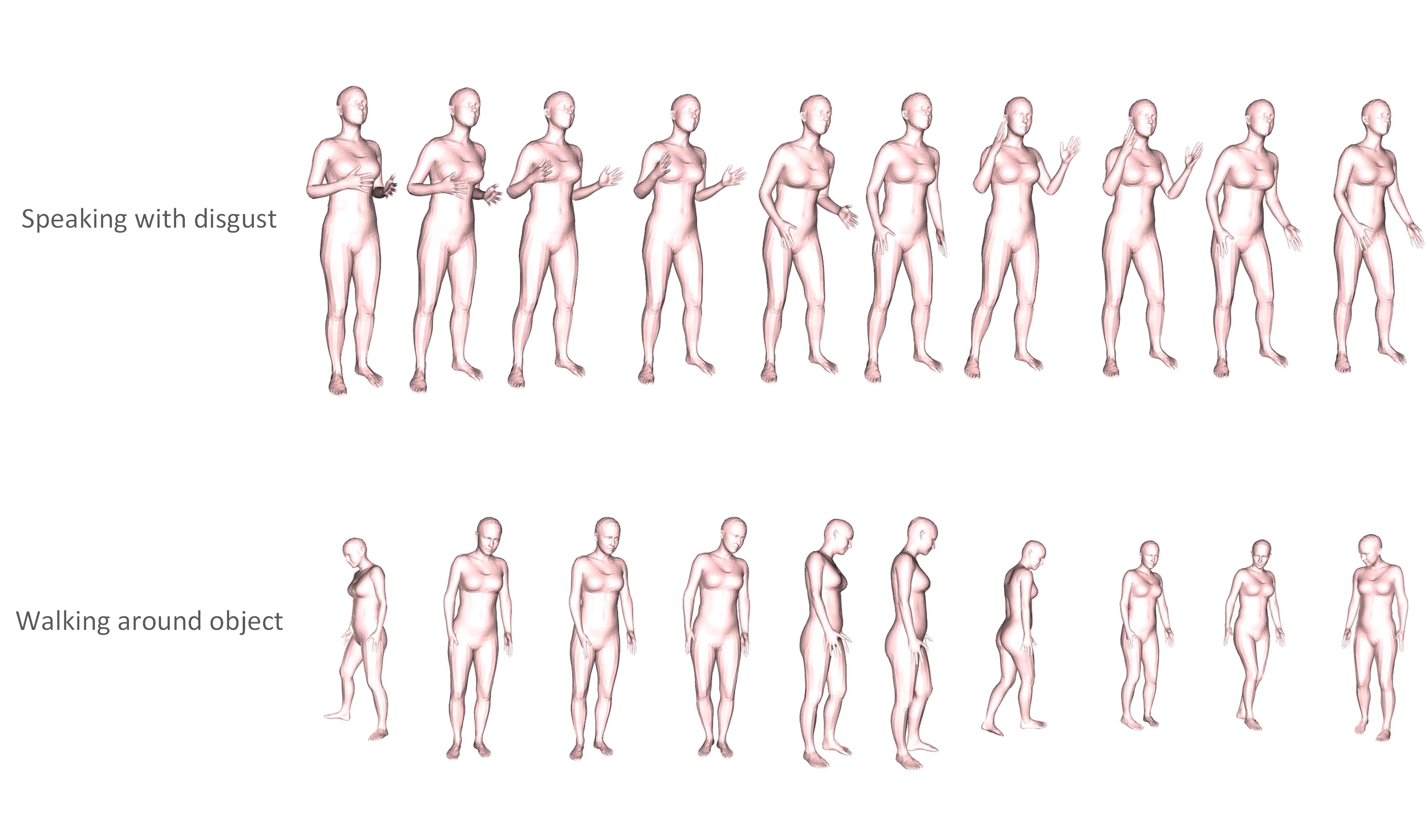}
    \caption{We train our framework on distinct sequences which present different actions and emotions. When conditioned on one-hot encodings, our model is able to synthesize variations of the training examples. For instances, we can generate motion of a person talking with disgust (top) or walking around an object (bottom).}
    \Description{.}
    \label{fig:disgust_walking}
\end{figure*}

\subsection{Training Data}
% start first with performance capture dta and such, 
We train and test our method on multiple sequences featuring a subject who describes objects and discusses different topics. We capture the subject's motion by employing a modified version of EasyMocap \cite{easymocap}, tailored to estimate SMPL \cite{loper2015smpl} parameters from performance capture. The subject's poses are characterized by 23 joints plus their global orientation and displacement. As detailed in Section \ref{sec:repr}, we incorporate contact labels for the feet, computed from three vertices on each foot --- heel, big toe, and small toe. This results in a total of 219 features per frame.

For the task of conditional motion synthesis, our training data has a total duration of 3 minutes and 40 seconds. The dataset comprises four sequences, each featuring an actress either describing an artwork or discussing past experiences with positive or negative sentiments. We assign a unique label to each training sequence, which is encoded as a one-hot vector for the conditioning process. This setup allows the model to learn motion generation conditioned on specific narrative contexts, as displayed in Figure \ref{fig:disgust_walking}.

\begin{figure*}[htb]
    \centering
    \includegraphics[width=\linewidth]{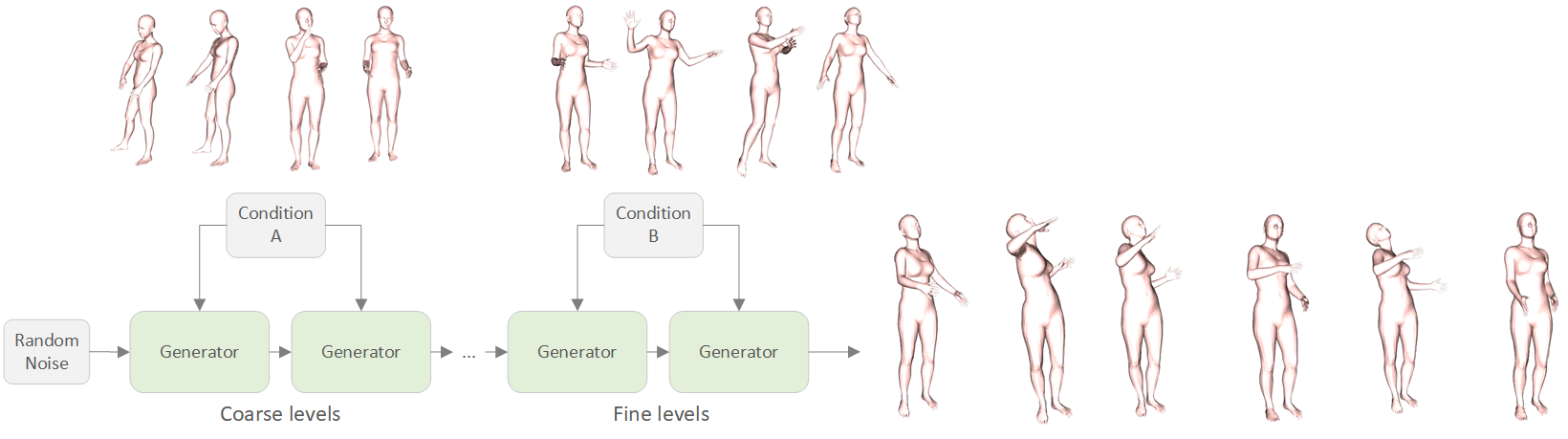}
    \caption{Multi-scale motion generation framework: Our approach learns embeddings across various temporal resolutions from a set of control signals, which enables the synthesis of diverse motions that integrate components from different example sequences or inputs. For example, our model can generate a sequence of a person describing an object (condition A) while incorporating the joyful speaking motion of another sequence (condition B).}
    \Description{.}
    \label{fig:mixing}
\end{figure*}

% ..............
\subsection{Implementation Details}

Our implementation follows the architecture outlined in \cite{li2022ganimator}, with adjustments to accommodate the specific motion channel dimensions and skeletal structure used in our experiments. We use the PyTorch framework for our model implementation. For optimization, we employ the Adam optimizer with a learning rate of $10^{-4}$, and we train the model for a different number of iterations depending on the task. The hyperparameters of the loss function (\ref{eq:loss}) are set as follows: $\lambda_{\text{adv}}=1$, $\lambda_{\text{rec}}=50$, $\lambda_{\text{con}}=5$, and $
\lambda_{\text{smooth}}=5$.

We design our generator to synthesize motion based on multiple conditions. Each condition, or control signal $s$, corresponds to a different training sequence, represented by a one-hot encoded vector. These one-hot vectors are embedded into a learned feature space with a size of 8.

During training, we aim to teach the network to generate motion sequences that can be controlled by different conditions at different resolution levels, as depicted in Figure \ref{fig:mixing}. Inspired by style mixing regularization techniques from \cite{karras2020analyzing}, we apply a similar strategy to our model. Specifically, 90\% of the generated motion sequences involve this regularization, where we use two conditions sampled randomly. A condition $s^{a}$ is applied to the motion in steps interval $[1, i]$, and condition $s^b$ is applied in steps interval $[i+1, L]$, with the crossover point $i$ being selected randomly. This approach encourages the network to disentangle and mix different conditions effectively across different resolution levels.

However, we avoid mixing these conditions when computing the reconstruction loss $\mathcal{L}_{\text{rec}}$, allowing the network to learn that a single condition should correspond to a single sequence. This training strategy helps the model blend different conditions across various resolution levels during inference, enabling the generation of motion sequences that combine coarse motion characteristics from one condition with fine-grained details from another. This method contrasts with GANimator \cite{li2022ganimator}, which trains separate models on individual sequences and combines them at different levels. Our approach achieves a similar goal but within a unified model, facilitating control over which resolution levels correspond to specific conditions.

Finally, we train for $15000$ iterations in the first two resolution levels and $25000$ iterations for the remaining resolution levels.

\subsection{Evaluation and Discussion}

We evaluate coverage and diversity of our model with common metrics for motion synthesis and compare it with an implementation of GANimator~\cite{li2022ganimator} adjusted to synthesize SMPL pose parameters.

While reconstruction-based metrics such as RMSE are commonly used for assessing the accuracy of generated motion compared to ground truth, they are less informative in the context of our work. Both our method and prior approaches are capable of nearly perfect reconstruction of motion sequences, leading to RMSE values close to zero. However, the primary goal of this work is not merely to reconstruct motion, but to generate novel variations of the training data, synthesizing new and diverse sequences. Therefore, we focus on more relevant metrics such as local and global diversity, as well as the coverage over the training set, which provide better insight into the model's ability to create varied and natural motion sequences. These metrics reflect the richness and variability of the generated motion rather than strict accuracy to a single reference. 

\subsubsection{Coverage} This metric evaluates the extent to which our model captures all temporal windows in the training sequences.

Since there is a limited number of training examples, we measure the coverage on all possible temporal windows of each example. More specifically, for a window $\mathcal{W}(\Theta, T') = \{\Theta^{t:t+T'-1}\}_{t=1}^{T-T'-1}$ of a given length $T'$, where $\Theta^{k:l}$ denotes the sequence of frames $k$ to $l$ of the training example $\Theta$, which has a total length $T$. Given a generated result $\hat{\Theta}$, we label a temporal window $\Theta_{\mathcal{W}}\in \mathcal{W}(\Theta, T'_c)$ as covered if its distance measure to the nearest neighbor in $\hat{\Theta}$ is smaller than an empirically chosen threshold $\epsilon$. The coverage of animation $\hat{\Theta}$ on $\Theta$ is defined as

\begin{equation}
    \text{Cov}(\hat{\Theta},\Theta)=\frac{1}{|\mathcal{W}(\Theta,T'_c)|}\sum_{\Theta_{\mathcal{W}}\in\mathcal{W}(\Theta,T'_c)}\mathbb{I}[\text{NN}(\Theta_{\mathcal{W}},\hat{\Theta})<\epsilon].
\end{equation}

$\text{NN}(\Theta_1,\Theta_2)$ denotes the distance of the nearest neighbor of the animation sequence $\Theta_1$ in $\Theta_2$. As distance measure we use the Frobenius norm on the local joint rotation matrices. We choose $T_c=30$, capturing local patch length of 1 second. Similarly, the coverage of mode $\mathcal{G}(\cdot)$ on $\Theta$ is defined by
\begin{equation}
    \text{Cov}(\mathcal{G},\Theta)=\mathbb{E}_z\text{Cov}(\mathcal{G}(z),\Theta).
\end{equation}
\subsubsection{Global diversity} This measure gauges the overall structural diversity in comparison to a training example by measuring the distance with patch nearest neighbors (PNN).

To quantitatively measure the global structure diversity against a single training example, we measure the distance between patched nearest neighbors (PNN). The idea is to divide the generated animation into several segments, where each segment is no longer than a threshold $T_{\text{min}}$, and find a segmentation that minimizes the average per-frame nearest neighbor cost. For each frame $t$ in the generated animation $\hat{\Theta}$ we match it with frame $l_t$ in the training animation $\Theta$, such that between every neighboring points in the set of discontinuous points $\{t|l_t\neq l_{t-1}+1\}$ is at least $T_{\text{min}}$. This is because every discontinuous point corresponds to the starting point of a new segment. We call such an assignment $\{l_t\}_{i=1}^{T'}$ a \textit{segmentation} on $\hat{\Theta}$. When a large global structure variation is present, it is difficult to find a close nearest neighbor for $T_{\text{min}}$. The patched nearest neighbor is defined by, minimizing over all possible segmentations:

\begin{equation}
    \mathcal{L}_{\text{PNN}} = \min_{\{l_t\}} \frac{1}{T} \sum_{t=1}^T \|\hat{\Theta}^t-\Theta^{l_t}\|_F^2,
\end{equation}

where $\hat{\Theta}^t$ denotes frame $t$ in $\hat{\Theta}$ and $\Theta^{l_t}$ denotes frame $l_t$ in $\Theta$. We choose $T_{\text{min}}=30$.

\subsubsection{Local diversity} This metric assesses the diversity of individual frames by comparing each local window $\hat{\Theta}_{\mathcal{W}}\in\mathcal{W}(\hat{\Theta},T_d)$ of length $T_d$ to its nearest neighbor in a training sequence $\Theta$.

\begin{equation}
    \mathcal{L}_{\text{local}} = \frac{1}{|\mathcal{W}(\hat{\Theta},T_d)|} \sum_{\hat{\Theta}_{\mathcal{W}}\in\mathcal{W}(\hat{\Theta},T_d)} \text{NN}(\hat{\Theta}_{\mathcal{W}},\Theta).
\end{equation}

Similar to the definition of coverage, we use the Frobenius norm over local joint rotations. We choose $T_d=15$ to capture local differences between the generated results and the training example.

We compute these metrics for each training sequence and present the results in \autoref{tab:comparison}, where sequences A-C include the performer describing and walking around an object, while sequences C and D feature the actress discussing past events in an angry and joyful manner. Notably, by incorporating a control signal, our model achieves higher coverage across each example sequence. Furthermore, leveraging the capability to blend content and details across various resolution levels, our model generates more diverse motions, as shown in the supplementary videos. This underscores the adaptability of our framework in producing a wide range of motion patterns.

\begin{figure}
    \centering
    \includegraphics[width=\linewidth]{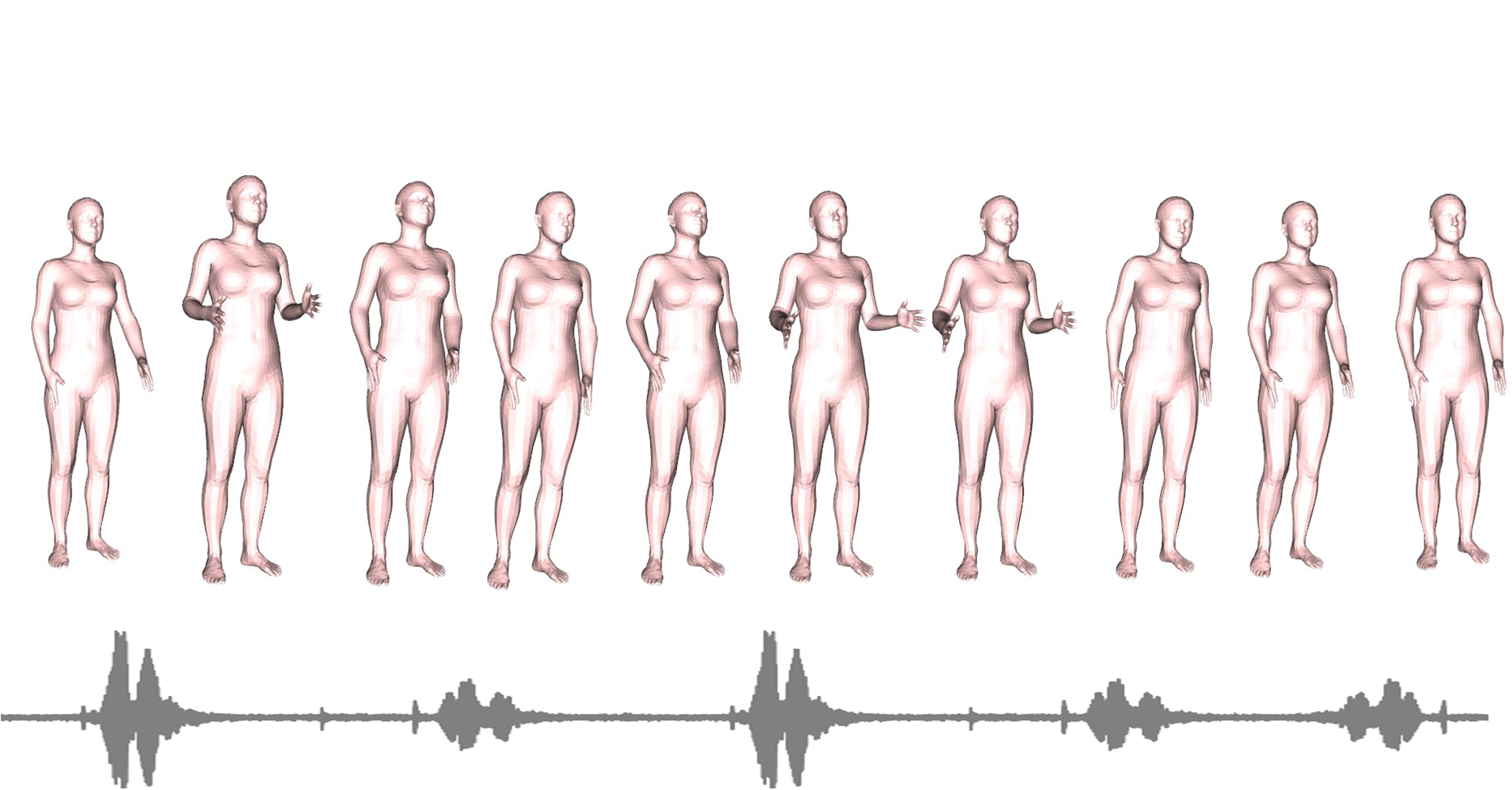}
    \caption{Our model is trained using paired speech features and motion data, with additional unpaired speech samples to enhance generalization. The result is the synthesis of co-speech gestures that are synchronized with the input speech.}
    \Description{.}
    \label{fig:gestures}
\end{figure}

\begin{figure*}[htb]
    \centering
    \includegraphics[width=\linewidth]{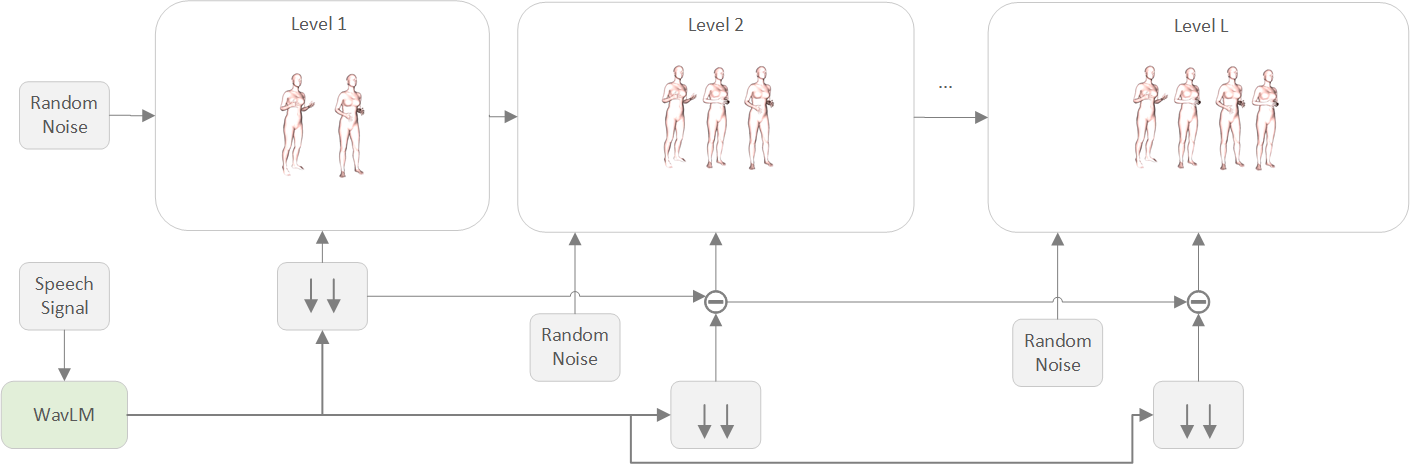}
    \caption{Gesture synthesis: Speech features are aligned with motion data by downsampling WavLM-extracted features from 50fps to 25fps and further downsampled to match the lower resolutions of our framework. Residual information is extracted for finer levels, ensuring that each level captures incremental motion details.}
    \Description{.}
    \label{fig:speech}
\end{figure*}

\section{Co-Speech Gestures from Limited Data}
\label{sec:appl}

We explore the capability of our method in a more complex scenario: synthesizing gestures from speech, particularly in situations with limited data. For this experiment, we utilize a dataset containing speech and motion data from three subjects, totaling 23 minutes and 1 second of recorded sequences. The dataset comprises 16 sequences, with each featuring a single subject independently discussing a piece of art or various topics. To augment the training data, we include an additional 27 unpaired audio files from the same subjects, which contain speech without corresponding motion sequences.

All paired and unpaired audio data are preprocessed by extracting features using WavLM \cite{chen2022wavlm}, a speech model designed to capture both acoustic and linguistic features. These features are then fed into our encoder networks $\{S_i\}_{i=1}^L$, which compute the corresponding FiLM parameters. These parameters modulate the upper body motion of the character, before the generator networks contribute missing details, enabling the synthesis of gesture sequences that are synchronized with the speech input.

%Since our focus is on the relationship between speech and gesture, rather than on the semantics of the spoken content, we include audio in both English and German within the same training set.

Furthermore, to align the resolution of the speech features with that of the motion data, we first interpolate the WavLM-extracted speech features from their original resolution of approximately 50fps down to the motion framerate of 25fps. We then continue to downsample both paired and unpaired audio samples to match the lower resolutions required by the various levels of our framework, constructing an initial set of features, denoted as $\{\tilde{s}_i\}_{i=1}^L$. Next, we transform these features to represent only the residual information for the finer resolution levels. Specifically, for levels $i > 1$, we compute $s_i = \tilde{s}_{i-1} - \tilde{s}_i$, while for the first level, we use $s_1 = \tilde{s}_1$. This residual representation ensures that the finer levels capture incremental details rather than redundant information. We illustrate this process in Figure \ref{fig:speech}.

To further enhance the network's ability to generalize, we employ paired speech and motion data during reconstruction steps, while unpaired audio samples are used for random generation. In other words, we train with the reconstruction loss (\ref{eq:recloss}) using paired data exclusively, while we optimize the network for random generation (without a reconstruction loss) using only unpaired speech samples. By alternating between random and reconstruction processes, the network learns the correlations between speech and movement while also developing the flexibility to generalize beyond the paired data, as illustrated in Figure \ref{fig:gestures}.

% we also load weights from previous levels to speed up training.

\section{Conclusion}
\label{sec:concl}

% Conclusion and future work.
In this work, we introduced a novel multi-resolution motion generation framework featuring style control, designed to excel with limited training data. Our approach utilizes skeletal convolution layers that model motion along the human kinematic chain, coupled with a multi-scale architecture that incrementally enhances motion details. In contrast to earlier approaches, our model provides enhanced control over both content and style in the generated motion, facilitated by our style modulation module. Additionally, by directly synthesizing sequences of SMPL pose parameters, we eliminate the need for test-time adjustments to fit the generated motion onto a human mesh.

We also explored the capability of our framework in the challenging task of gesture synthesis from speech, particularly under conditions of limited data. By integrating paired speech-motion data with unpaired speech samples, our model was able to generate co-speech gestures that are synchronized with the input speech. This demonstrates the flexibility and robustness of our approach in handling more complex scenarios, where the relationship between audio and motion must be captured and reproduced with high fidelity.

In the future, we aim at exploring conditional synthesis using alternative control signals, such as semantic features extracted from text. Furthermore, we plan to incorporate hand motion modeling, which is essential for achieving realistic human animation. Expanding the framework to include finer details like hand gestures will further enhance the realism and applicability of our model in diverse animation tasks.

\begin{acks}
\label{sec:acknowledgements}
This research has partly been funded by the European Union’s Horizon Europe research and innovation programme (Luminous, grant no.\ 101135724, SPIRIT, grant no. 101070672), and the German Ministry of Education and Research (VoluProf, grant no.\ 16SV8705).
\end{acks}
% \begin{acks}
% \end{acks}

\bibliographystyle{ACM-Reference-Format}
\bibliography{refs}

% \appendix









\end{document}